\documentclass{article}

    \usepackage[preprint, nonatbib]{neurips_2020}

\usepackage[utf8]{inputenc} 
\usepackage[T1]{fontenc}    
\usepackage{hyperref}       
\usepackage{url}            
\usepackage{booktabs}       
\usepackage{amsfonts}       
\usepackage{nicefrac}       
\usepackage{microtype}      
\usepackage{amsmath}        
\usepackage{graphicx}       

\title{MLPs to Find Extrema of Functionals}

\author{%
  Liu, Tao \\
}

\begin{document}

\maketitle

\begin{abstract}
Multilayer perceptron (MLP) is a class of networks composed of multiple layers of perceptrons, and it is essentially a mathematical function.
Based on MLP, we develop a new numerical method to find the extrema of functionals. As demonstrations, we present our solutions in three physic scenes. Ideally, the same method is applicable to any cases where the objective curve/surface can be fitted by second-order differentiable functions. This method can also be extended to cases where there are a finite number of non-differentiable (but continuous) points/surfaces. 
\end{abstract}

\section{Introduction}

If you are a creator who wants to create a regular world with the simplest of principles, then the principle of least action is probably your best bet, because it runs through classical physics to modern physics, including optics, electromagnetics, thermodynamics, hydromechanics, special and general relativity, quantum mechanics, and even string theory.

As early as in the 16th century, the Dutch physicist Snell discovered the law of refraction, and in the 17th century, the French mathematician Fermat further proposed the Fermat principle, which he called the principle of minimum time. The content is: light always chooses the path that takes the least time to travel. Although the Fermat principle applies only to geometric optics, it also plays a guiding role in other fields. For example, John Bernoulli once used this idea to successfully solve the problem of brachistochrone. With the idea of Fermat principle, and advanced by many mathematicians and physicists, the principle of least action was developed in the 18th century.

In physics, action is a functional which takes the path of the system as its argument and has a real number as its result. The principle of least action states that the path taken by the system is the one for which the action is stationary to first order (usually minimum). With variational method, the differential equation(s) that the minimum function satisfies can be derived. If instead, a differential equation was given first, sometimes we could obtain an equivalent form of functional minimum, according to the works of American mathematician Friedrichs\cite{friedrichs1958symmetric}.

In the present work, we are concerned with MLPs as a one-stop numerical solution to find extrema of functionals. Firstly, we argue that the action functional is the physical version of the so-called cost function(al) of MLP in Machine Learning and Optimal Control. Secondly, it is easy to optimize MLPs with the gradient-based algorithm. Thirdly, given that MLPs are universal function approximators as shown by Cybenko's theorem\cite{cybenko1989approximations}, they are potential to approximate any extremal functions.

Unlike the Rayleigh–Ritz method, which is a widely-used numerical method to solve extrema of functionals, the MLP method needs no approximating functions. This may make MLPs more advantageous to deal with problems with complex boundaries.

To illustrate this scheme, in the following paper, we will present MLP solutions to several physical problems, including movements in gravity field, refractions of light, and minimal surfaces, which are deliberately chosen to cover cases of different boundaries, joints, and dimensions. Before that, it is a short theoretical part.

\section{Theory}

The starting point is the action, denoted $\mathcal{S}$, of a physical system. It is technically a functional of the path  which defines the configuration of the system:
\begin{equation}
    \mathcal{S}[y(x)]=\int_{\mathbf{x}_1}^{\mathbf{x}_2}L(x, y(x), y'(x))dx
\end{equation} 
where $L(x, y(x), y'(x))$ is the Lagrangian. $y(x)$ is specified at two end points where $y(\mathbf{x}_1)=\mathbf{y}_1$ and $y(\mathbf{x}_2)=\mathbf{y}_2$. The principle of least action states that among all conceivable paths y(x) that could connect the given points, the true paths are those that make $\mathcal{S}$ stationary. In this paper, we only deal with situations where $\mathcal{S}$ is minimal\footnote{The method of Lagrange multiplier may help to find all stationary points. However, it will bring an explosion in calculations.}.

We now model the principle in the language of MLP :
\begin{equation}
\begin{cases}
   \min_{Y} \left[ \frac{(\mathbf{x}_2-\mathbf{x}_1)}{N}\sum_{i}^{N} L(x_i, Y(x_i), Y'(x_i))
    + \sum_{i}^{2} F(\mathbf{x}_i)  (Y(\mathbf{x}_i)-\mathbf{y}_i) \right]
\\ 
   \max_{F} \left[ \sum_{i}^{2} F(\mathbf{x}_i)  (Y(\mathbf{x}_i)-\mathbf{y}_i) \right]
\end{cases}
\end{equation}
The path function $y(x)$ is modeled using a second-order differentiable MLP ($Y$). The computation of integral, $\int_{\mathbf{x}_1}^{\mathbf{x}_2}$, is modeled by a summation of $N$ samples ($x_i$) sampled from the interval of $(\mathbf{x}_1, \mathbf{x}_2)$. The boundary conditions are modeled using a learning model ($F$), which dynamically regulates the power of (generalized) forces to bring $Y(\mathbf{x}_i)$ near $\mathbf{y}_i$. The models $Y$ and $F$ are opponents in the minimax gaming. If with good learning rate and dampening strategy, the system will come to an equilibrium state where $Y(\mathbf{x}_i) \rightarrow  \mathbf{y}_i$.

\section{Demos}
\subsection{Movement in the gravity field}
\begin{figure}[htb]
  \centering
  \includegraphics[width=\textwidth]{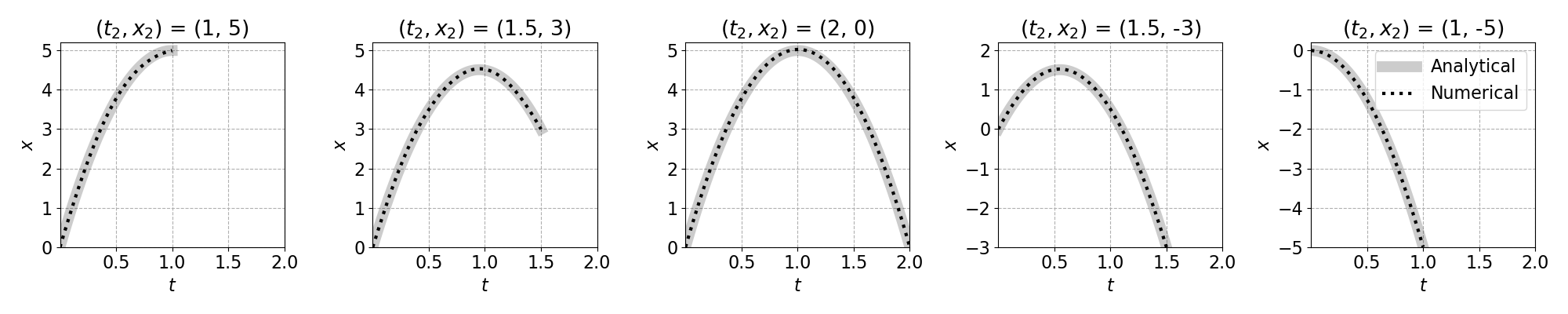}
  \caption{\label{fig:gravity} Movement in the gravity field numerically emulated by MLPs}
\end{figure}

Suppose there is a particle in the gravitational field, which is thrown out, then goes up and comes down. The Lagrangian here is $L(t,x(t), \dot{x}(t)) = \frac{1}{2}m\dot{x}^2 - mgx$, where we will set $m=1$ and $g=10$. The particle goes from an original place to a final place in a certain amount of time, and the boundary condition is  $x(\mathbf{t}_1)=\mathbf{x}_1$ and $x(\mathbf{t}_2)=\mathbf{x}_2$. The numerical emulations given by MLPs are shown in Figure~\ref{fig:gravity}.

\subsection{Refraction of light}
\begin{figure}[htb]
  \centering
  \includegraphics[width=\textwidth]{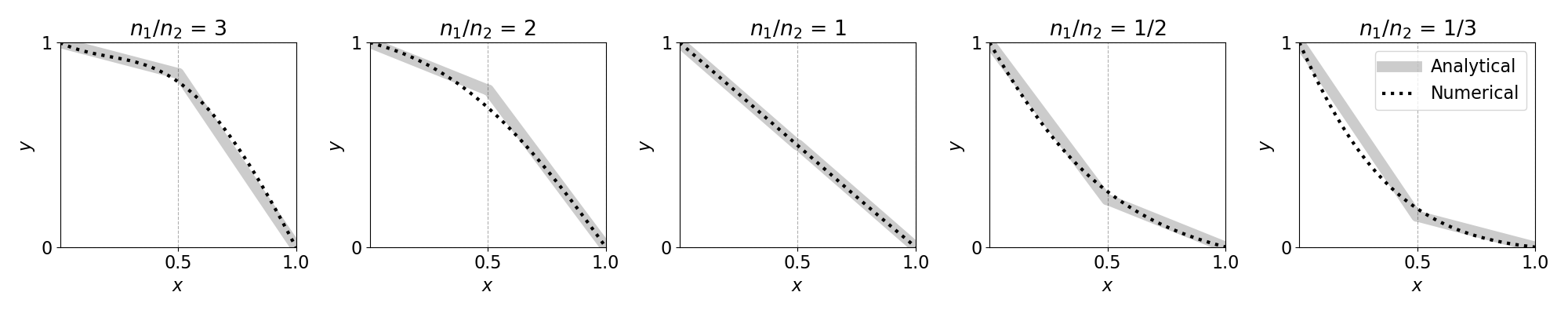}
  \caption{\label{fig:opt0} Refraction of light numerically emulated by MLPs}
\end{figure}

Fermat's principle states that the path taken by a ray between two given points is the path that can be traversed in the least time, and it was proposed as a means of explaining the ordinary law of refraction of light. The Lagrangian in propagation of light is $L(x, y(x), y'(x))=\frac{n}{c}\sqrt{1+ (y')^2}$, where n is the refractive index. 
Suppose that light passes through two media from left to right. The refractive indexes are $n_1$ and $n_2$ respectively. The numerical emulations performed by MLPs are shown in Figure~\ref{fig:opt0}.

The result reveals a flaw that only one network cannot fit angular lines well, because the network has to be second-order differentiable. In this case, we will couple two MLPs ($Y_1$ and $Y_2$) to model the left and right parts respectively. And at their interface, the paths will be jointed by a pair of forces in the same magnitude and opposite directions:
\begin{equation}
    \begin{cases}
   \min_{Y_1} \left[ \frac{(\mathbf{x}_1-\mathbf{x}_0)}{N}\sum_{i}^{N} L_1
    +  F(\mathbf{x}_0)  (Y_1(\mathbf{x}_0)-\mathbf{y}_0) 
    +  F(\mathbf{x}_1)  (Y_1(\mathbf{x}_1)-Y_2(\mathbf{x}_1)) 
\right]
\\
   \min_{Y_2} \left[ \frac{(\mathbf{x}_2-\mathbf{x}_1)}{N}\sum_{i}^{N} L_2
    +  F(\mathbf{x}_2)  (Y_2(\mathbf{x}_2)-\mathbf{y}_2) 
    +  F(\mathbf{x}_1)  (Y_1(\mathbf{x}_1)-Y_2(\mathbf{x}_1)) 
\right]
\\ 
   \max_{F} \left[F(\mathbf{x}_0)  (Y_1(\mathbf{x}_0)-\mathbf{y}_0) 
   + F(\mathbf{x}_1)  (Y_1(\mathbf{x}_1)-Y_2(\mathbf{x}_1))
   + F(\mathbf{x}_2)  (Y_2(\mathbf{x}_2)-\mathbf{y}_2) 
\right]
\end{cases}
\end{equation}

\begin{figure}[htb]
  \centering
  \includegraphics[width=\textwidth]{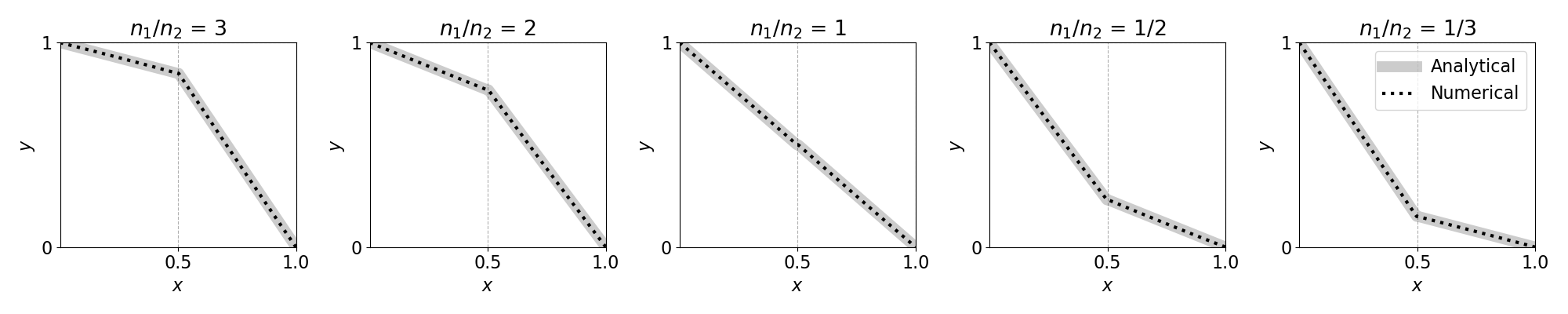}
  \caption{\label{fig:opt} Refraction of light numerically emulated by jointed MLPs}
\end{figure}

In this way, the numerical emulations performed by MLPs are shown in Figure~\ref{fig:opt}.

\subsection{Minimal surface}
The problem of minimal surface is to find the minimal surface of a boundary with specified constraints, which is known as Plateau problem. Here, the boundary is a given curve in three dimensions. Suppose that there is a soap film attached to a circle with a radius of 1. It has a surface tension coefficient $\sigma$ and an internal/external pressure difference $p$. The Lagrangian of this system is $L(x, y, z, \frac{dz}{dx}, \frac{dz}{dy}) = 2\sigma \sqrt{1 + (\frac{dz}{dx})^2 + (\frac{dz}{dy})^2} -p\cdot z$.

\begin{figure}[htb]
  \centering
  \includegraphics[width=\textwidth]{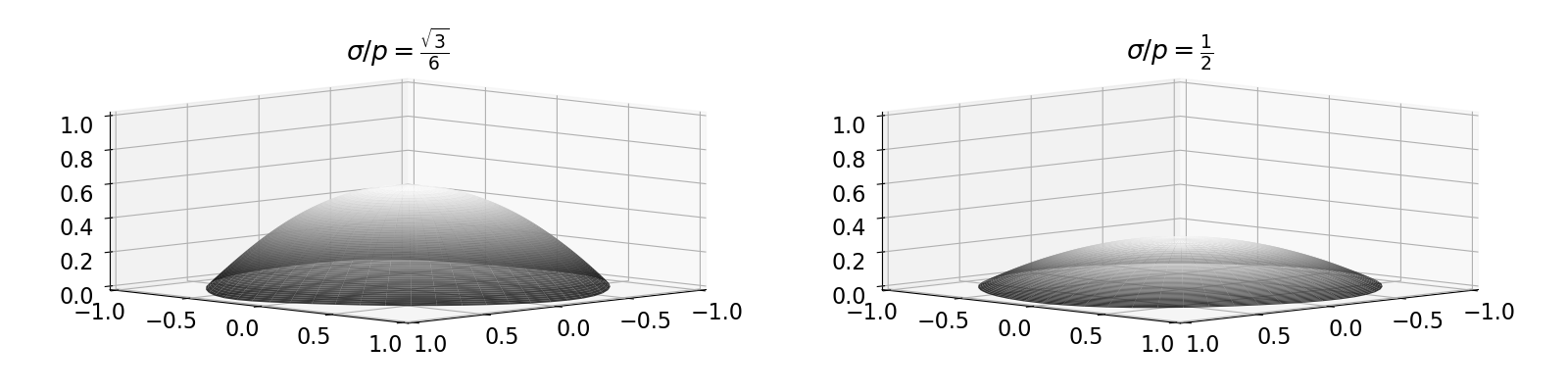}
  \caption{\label{fig:surface} Minimal surface of soap films numerically emulated by two-dimensional MLPs}
\end{figure}

Because the boundary is continuous, in our numerical emulations, the boundary force $F$ will be modeled by an MLP, too. The results are shown in Figure~\ref{fig:surface}.

\section{Conclusion}
In this paper, we introduce a physical term, namely Action, into MLP. The Action is technically a functional which maps from the MLP itself to a scalar quantity.
Then we use gradient-based optimization of MLPs as a numerical method to approximate the extremal function minimizing the value of action.
As an application demonstration, we emulate a few physical systems which follow the principle of least action.
Significantly, our approach requires no experimental data, only equations of actions and boundary conditions.

The purpose of this paper is only to show the feasibility, and does not include a serious evaluation of accuracy. We preliminarily think that a small batch size will cause a decrease in precision.

\bibliographystyle{apalike}
\small
\bibliography{reference}
\end{document}